\documentclass[10pt,twocolumn,superscriptaddress,floats,showpacs,nobalancelastpage,longbibliography,prb]{revtex4-2}

\usepackage[pdftex]{color}
\usepackage[dvipsnames]{xcolor}
\usepackage{braket}

\usepackage{array,longtable}
\newcolumntype{L}{>{\tiny $}p{0.33\columnwidth}<{$}}
\newcolumntype{M}{>{\scriptsize $}p{0.33\columnwidth}<{$}}
\newcolumntype{N}{>{\scriptsize $}p{0.43\columnwidth}<{$}}
\usepackage{amsmath}
\usepackage{amssymb}
\usepackage{amsfonts}
\usepackage{graphicx}
\usepackage{hyperref}
\usepackage{tabularx}
\usepackage{wasysym}
\usepackage{bm}

\usepackage{float}
\usepackage{graphics}
\usepackage[caption=false]{subfig}
\usepackage{tikz}
\usepackage{times}
\usepackage{dsfont}
\usepackage{subfig}

\usepackage[normalem]{ulem}

\allowdisplaybreaks[1]

\newif\ifhyper
\hypertrue
\ifhyper
\hypersetup{
   citecolor = {OliveGreen},
   colorlinks = {true}, 
   urlcolor = {RoyalBlue}, 
   linkcolor = {RoyalBlue}
}
\fi

\begin{document}

\title{Tensor‑Augmented Convolutional Neural Networks: \\ Enhancing Expressivity with Generic Tensor Kernels}

\author{Chia-Wei Hsing}
\email{cwhsing0219@gmail.com}
\affiliation{blueqat Inc., 2-24-12-39F, Shibuya, Shibuya-ku, Tokyo 150-6139, Japan}
\affiliation{Center for Quantum Science and Engineering, National Taiwan University, Taipei 10617, Taiwan}

\author{Wei-Lin Tu}
\email{weilintu@keio.jp}
\affiliation{Graduate School of Science and Technology, Keio University, Yokohama, Kanagawa 223-8522, Japan}
\affiliation{Keio University Sustainable Quantum Artificial Intelligence Center (KSQAIC), Keio University, Tokyo 108-8345, Japan}


\begin{abstract}

Convolutional Neural Networks (CNNs) excel at extracting local features hierarchically, but their performance in capturing complex correlations hinges heavily on deep architectures, which are usually computationally demanding and difficult to interpret.
%
To address these issues, we propose a physically-guided shallow model: tensor‑augmented CNN (TACNN), which replaces conventional convolution kernels with generic tensors to enhance representational capacity.
%
This choice is motivated by the fact that an order-$N$ tensor naturally encodes an arbitrary quantum superposition state in the Hilbert space of dimension $d^N$, where $d$ is the local physical dimension, thus offering substantially richer expressivity.
Furthermore, in our design the convolution output of each layer becomes a multilinear form capable of capturing high-order feature correlations, thereby equipping a shallow multilayer architecture with an expressive power competitive to that of deep CNNs.
%
On the Fashion-MNIST benchmark, TACNN demonstrates clear advantages over conventional CNNs, achieving remarkable accuracies with only a few layers. In particular, a TACNN with only two convolution layers attains a test accuracy of 93.7$\%$, surpassing or matching considerably deeper models such as VGG‑16 (93.5$\%$) and GoogLeNet (93.7$\%$).
%
These findings highlight TACNN as a promising framework that strengthens model expressivity while preserving architectural simplicity, paving the way towards more interpretable and efficient deep learning models.

\end{abstract}

\maketitle

\section{Introduction}
\label{sec:introduction}
 
Convolutional Neural Networks (CNNs) have emerged as a foundational architecture in deep learning, particularly for tasks involving structured data such as images, audio, and time series~\cite{Zhao2024}. 
This architectural paradigm has enabled significant advances in computer vision (CV), including image classification, object detection, semantic segmentation, and video analysis. Their ability to capture local features and hierarchical patterns has also made them increasingly relevant for analyzing lattice systems, spin models, and quantum phase transitions~\cite{Carrasquilla2021, Vicentini2022}. 
Despite these successes, conventional CNNs often require deep and computationally intensive architectures to achieve high accuracy, especially when modeling systems with complex correlations. 
As modern applications demand both higher predictive performance and better interpretability, there is a pressing need for new approaches that can enhance the representational power of CNNs without relying heavily on excessive depth or parameter count.
%
%

Recent efforts have sought to improve CV performance by incorporating ideas inspired by tensor‑based representations~\cite{Orus2019, Cirac2021, Banuls2023}.
CNNs and tensor‑network (TN) models offer distinct yet complementary perspectives on data representation. 
CNNs process inputs through layers of convolution kernels, producing hierarchical feature maps that effectively capture local patterns. This coarse‑graining mechanism enables the extraction of spatially localized features and underpins their success across a wide range of deep learning tasks. 
In contrast, TN approaches are designed to capture long‑range correlations, a capability essential for describing the intricate behavior of quantum many‑body systems. By tuning the bond dimension, the size of the indices connecting individual tensors, the expressive power of these models can be systematically increased, making them powerful ansätze for complex quantum states. 
Motivated by these strengths, recent research has explored the adaptability of TN‑inspired architectures to machine learning (ML), producing decent benchmarking results and extending their applicability well beyond traditional physics‑oriented problems~\cite{Novikov2015, Cohen2016, stoudenmire2016, Stoudenmire2018, Han2018, Glasser2018, Levine2019, Liu2019, Glasser2019, Efthymiou2019, Arbel2020, Cheng2021, Wang2023, Meng2023, Chen2024, Nie2025}.

However, despite extensive efforts to apply various TN architectures to machine-learning tasks, their performance remains largely limited. For example, in image classification on small benchmark datasets, their accuracies are notably below the state-of-the-art ones achieved by deep CNNs~\cite{Garcia2024}, especially on the more challenging Fashion-MNIST dataset~\cite{Chen2024, Nie2025, Meshkini2020}, even though in some cases TN models have comparable or more variational parameters. 
This disparity highlights a fundamental difference in representational priorities: while TN approaches are designed to capture long-range quantum correlations, which are essential for modeling entangled many-body systems, such features may not be prominently encoded in classical data, which are often dominated by statistical regularities and local patterns.
From a physical standpoint, it suggests that these quantum-inspired structures, while theoretically rich, may not align with the inductive biases required for optimal performance on classical data distributions. 
In contrast, CNNs excel at extracting local features through hierarchical convolution operations, and empirical evidence shows that modestly increasing the kernel size can significantly enhance model performance. 
This observation points to the critical role of local correlations in data-driven modeling and implies that effectively capturing and interpreting these localized structures may be more important than modeling global entanglement, at least within the scope of standard machine-learning benchmarks. 
As such, designing architectures that prioritize and maximize local expressivity while maintaining computational efficiency could be the key to bridging the gap between physically motivated models and practical artificial-intelligence (AI) systems.

A further key insight arises from examining how feature correlations are handled in conventional CNNs. Although increasing the number of convolution kernels typically improves model accuracy, the correlations among these kernels remain implicit and inaccessible during training. 
Still, because different kernels learn complementary aspects of the data, the presence of underlying correlations is undeniable. In contrast, a generic tensor naturally expands the state vector within the full Hilbert space, intertwining product components in a structured and expressive manner. 
By replacing standard kernels with such tensors, the model naturally embeds the aforementioned correlations directly into each kernel. As a result, even a small tensor, interpretable as a compact small-sized quantum state, can capture richer local structure, enabling the network to represent complex correlations with far fewer kernels. 
This provides both a conceptual and practical advantage, offering a more principled way to encode interactions that CNNs otherwise learn only implicitly.

Motivated by these considerations, we introduce a new strategy for enhancing convolutional architectures: embedding small quantum‑state structures directly into convolution kernels.
%
%
%
%
This leads to the proposed tensor‑augmented CNN (TACNN), in which each kernel is replaced by a fully generic higher‑order tensor. Such tensors naturally encode arbitrary quantum superposition states, providing substantially greater expressive power than conventional kernels while preserving the simplicity and scalability of the CNN framework. 
As demonstrated later, TACNN achieves consistently strong performance in the classical image‑classification task on the Fashion‑MNIST dataset~\cite{xiao2017fashionmnist}.
In particular, our benchmarking with TACNN shows results achieving a top accuracy of 93.7$\%$ on Fashion‑MNIST, surpassing or matching the performance of much deeper architectures such as VGG‑16 (93.5$\%$) and GoogLeNet (93.7$\%$)~\cite{xiao2017fashionmnist}. 
Although we benchmark TACNN primarily with image classification, the underlying principle of enhancing convolution operators through quantum-inspired tensorization is general and can be readily extended to a broad range of ML tasks involving structured or correlated data. We believe that this combination of enhanced expressivity, efficiency, interpretability, and broad applicability positions TACNN as a conceptually novel and practically powerful framework for advancing explainable deep learning.
\section{Architectures and Theories}
\label{sec:architectures}
%
%
\subsection{Single-layer TACNN}
\label{subsubsec:singletacnn}
\begin{figure}[tbp]
\centering
\includegraphics[width=1.0\linewidth]{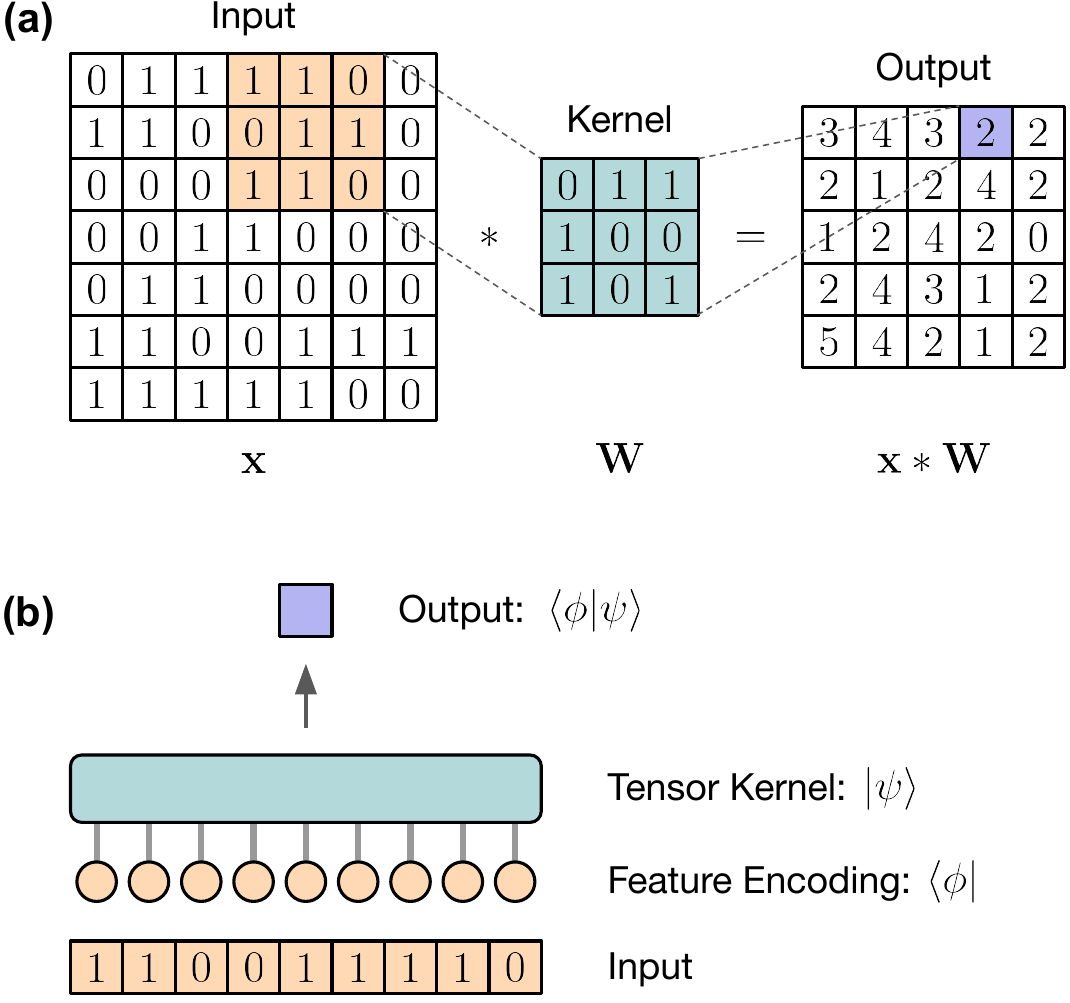}
\caption{Convolution operation in (a) a conventional CNN and (b) the proposed TACNN. 
(a) In a standard CNN, an element-wise multiplication is performed between each local image patch and a convolution kernel, and a following summation produces a scalar output. This is equivalent to an inner product between two flattened arrays.
(b) In TACNN, the input patch is first mapped into a higher‑dimensional Hilbert space, forming a product state $\langle \phi|$, and each convolution kernel is replaced by a generic tensor representing an arbitrary superposition state $|\psi\rangle$. Similarly, the inner product $\langle \phi|\psi\rangle$  yields a scalar, yet with substantially enhanced expressive capacity as elaborated in Section~\ref{sec:architectures}.}
\label{Fig1}
\end{figure}
%
In the standard CNN protocol, a labeled image comprises multiple input channels (3 for RGB), each of which corresponds to a grayscale image represented as $\mathbf{x}\in\mathbb{R}_{\geq 0}^{H\times W}$ with each and every pixel value $x\in\left[0,255\right]$, where $H\times W$ denotes the spatial resolution. Whereas conventional CNNs typically extract features linearly, recent studies on quantum-inspired methods have demonstrated that two-dimensional (2-d) feature extraction can be alternatively implemented by embedding the input into a Hilbert space of higher-dimension~\cite{Cheng2021}.
It has been shown that there are several options of feature encoding commonly used in TN-based machine-learning models. For better performance and explainability of the proposed TACNN, we choose a mapping $f:x\mapsto[x, 1-x]$. More specifically, each normalized pixel value $x(\mathbf{R})\in\left[0,1\right]$ is mapped through a feature‑encoding function $f$ to a $2$-d vector:
\begin{equation}
\ket{x(\mathbf{R})} = x(\mathbf{R})\ket{0} + (1-x(\mathbf{R}))\ket{1}
= \begin{pmatrix} x(\mathbf{R})\\1-x(\mathbf{R}) \end{pmatrix},
\label{Eq1}
\end{equation}
where
\begin{equation}
\ket{0} = \begin{pmatrix} 1\\0 \end{pmatrix}
\quad \text{and} \quad
\ket{1} = \begin{pmatrix} 0\\1 \end{pmatrix}
\label{Eq2}
\end{equation}
denote the white and black states, respectively.
Here we define $\mathbf{R}=(\mathbf{r}, \mathbf{k})$, where $\mathbf{r}$ denotes the position of a patch within the $H\times W$ grid and $\mathbf{k}$ labels each pixel within the corresponding patch.
%
%
For a local patch with $N$ pixels, the patch state $\ket{\phi(\mathbf{r})}$ is represented by the tensor product state
\begin{equation}
\ket{\phi(\mathbf{r})} = \bigotimes_{k=1}^{N}\ket{x(\mathbf{R})} \; \in (\mathbb{R}^2)^{\otimes N}
\label{Eq3}
\end{equation}
in the Hilbert space of dimension $2^N$. 
%

%
Following the standard design principles of CNNs, we employ multiple convolution kernels to capture the diverse local features in an image. 
However, unlike conventional CNNs, where each kernel is represented by a simple $L\times L$ array as schematically shown in Fig.~\ref{Fig1}(a), the convolution kernels in TACNN are constructed with generic tensors of higher order, as illustrated in Fig.~\ref{Fig1}(b). The total number of such tensor-based convolution kernels is denoted by $N_\text{TK}$. 
%
%
For a single convolution layer, an order-$N$ tensor kernel can be expressed as a general superposed state 
\begin{equation}
\ket{\psi_{j}} = \sum \limits_{\mathbf{s}} {c_{j}({\mathbf{s}})} \ket{\mathbf{s}} \;\; \text{with} \;\; \mathbf{s}=({s_1},{s_2},\ldots{s_{N}}),
\label{Eq4}
\end{equation}
where $j$ denotes the kernel index running from $1$ to $N_\text{TK}$, the amplitudes $c_{j}({\mathbf{s}}) \in \mathbb{R}$ are the trainable parameters, and the configuration $\mathbf{s}\in\{0,1\}^N$. 
In contrast to the convolution kernels in conventional CNNs, where each of which encodes a single pattern, every tensor kernel in TACNN represents a coherent superposition over all $2^N$ binary configurations. This equips each kernel with an exponentially larger expressive capacity, laying the first theoretical foundation for the potential advantage of TACNN.
With the above definition, the convolution operation is thereby formulated as the inner product between the patch state and the kernel state:
\begin{equation}
\begin{aligned}
y_{j}(\mathbf{r}) &= \braket{\phi(\mathbf{r})|\psi_{j}} \\ &= \sum\limits_{\mathbf{s}} {c_{j}({\mathbf{s}})} \prod\limits_{k=1}^{N} x(\mathbf{R})^{1-s_k} (1-x(\mathbf{R}))^{s_k},
\end{aligned}
\label{Eq5}
\end{equation}
which is a multilinear form in $x(\mathbf{R})$ of each pixel within the patch. 
%
Although the trainable model parameters enter linearly through the coefficients $c_{j}({\mathbf{s}})$, the resulting input-output mapping departs significantly from linearity because of the multiplicative structure inherent in the basis function
\begin{equation}
\beta(\mathbf{r}, \mathbf{s}) = \prod \limits_{k=1}^{N} x(\mathbf{R})^{1-s_k} (1-x(\mathbf{R}))^{s_k}.
\label{Eq6}
\end{equation}
%
Each term in Eq.~(\ref{Eq5}) represents a basis $\beta(\mathbf{r}, \mathbf{s})$ of a specific many‑body configuration $\mathbf{s}$ weighted by $c_{j}({\mathbf{s}})$, and the output is obtained by linearly combining all the $2^N$ multilinear basis functions.
%
Thus, even a single tensor kernel induces a response that cannot be described by a linear mapping of the input, offering an expressive power beyond that resulting from a conventional convolution. This inherent multilinear structure is central to the ability of TACNN to capture high‑order features in an image.
%
Furthermore, in standard CNNs, since the result of a convolution operation is linear in the pixel values, additional layers with activation functions are required for the model to depart from linearity and capture higher‑order correlations. Hence, the per-layer expressivity of TACNN is significantly richer than that of CNNs. Together, these form the second theoretical foundation for the potential advantage of TACNN.
%
%

\subsection{Multilayer TACNN}
\label{subsubsec:Multitacnn}
In order to expand TACNN to a multilayer form, it is essential to pre-process the output from each layer properly.
For each output $y^{n}_{j_{n}}(\mathbf{r}_{n})$ of layer $n$ that enters the input channel $j_{n}$ of the subsequent layer $n'=n+1$, we apply the following mapping:
\begin{equation}
z^{n'}_{j_{n}}(\mathbf{r}_{n'}, \mathbf{k}_{n'}) = \sigma \left( \frac{ y^{n}_{j_{n}}(\mathbf{r}_{n}) - \bar{y}^{n}_{j_{n}}}{\text{std}(y^{n}_{j_{n}})}\right),
\label{Eq7}
\end{equation}
where $\bar{y}^{n}_{j_{n}}$ and $\text{std}(y^{n}_{j_{n}})$ represent the mean value and standard deviation of $y^{n}_{j_{n}}(\mathbf{r}_{n})$ calculated over all $\mathbf{r}_{n}$, respectively.
%
%
Note that we now attach a sub‑index to every function and variable to indicate the layer to which they belong.
In Eq.~(\ref{Eq7}), $\sigma (\cdot)$ denotes the sigmoid function, and the newly defined $\mathbf{R}_{n'}=(\mathbf{r}_{n'}, \mathbf{k}_{n'})$ is determined by a corresponding mapping from $\mathbf{r}_{n}$.
%
The smooth normalization scheme in Eq.~(\ref{Eq7}) ensures that the input of layer $n'$ follows
\begin{equation}
z^{n'}_{j_{n}}(\mathbf{R}_{n'}) \in \left[0,1\right], \; \forall \ j_{n} \wedge \mathbf{R}_{n'},
\label{Eq8}
\end{equation}
where the sigmoid function not only helps keep information loss minimal, but introduces nonlinearity for each layer.
By applying the same feature-encoding process as in Eq.~(\ref{Eq1}) and transforming each patch into a tensor product state in the same fashion as Eq.~(\ref{Eq3}), for layer $n'$ the patch state of input channel $j_{n}$ becomes
\begin{equation}
\ket{\phi^{n'}_{j_{n}}(\mathbf{r}_{n'})} = \bigotimes_{k_{n'}=1}^{N_{n'}}\ket{z^{n'}_{j_{n}}(\mathbf{R}_{n'})},
\label{Eq9}
\end{equation}
and the tensor kernels follow the general form of
\begin{equation}
\ket{\psi^{n'}_{j_{n},\ j_{n'}}} = \sum \limits_{\mathbf{s}_{n'}} {c_{j_{n},\ j_{n'}}^{n'}({\mathbf{s}}_{n'})} \ket{\mathbf{s}_{n'}}, \; {c_{j_{n},\ j_{n'}}^{n'}({\mathbf{s}}_{n'})} \in \mathbb{R},
\label{Eq10}
\end{equation}
where $j_{n'}$ denotes the index of the output channel, leading to a total of $N^{n}_\text{TK} \times N^{n'}_\text{TK}$ kernels for layer $n'$.
As with the single-layer case, the amplitudes $c_{j_{n},\ j_{n'}}^{n'}({\mathbf{s}_{n'}})$ encode the trainable parameters. 
%
%
The output of channel $j_{n'}$ now becomes:
\begin{equation}
\begin{aligned}
y^{n'}_{j_{n'}}(\mathbf{r}_{n'}) &= \sum \limits_{j_{n}=1}^{N^{n}_\text{TK}} \braket{\phi^{n'}_{j_{n}}(\mathbf{r}_{n'})) | \psi^{n'}_{j_{n},\ j_{n'}}} \\ &= 
\sum \limits_{j_{n}=1}^{N^{n}_\text{TK}} \sum \limits_{\mathbf{s}_{n'}} {c_{j_{n},\ j_{n'}}^{n'}({\mathbf{s}}_{n'})} \beta^{n'}_{j_{n}}(\mathbf{r}_{n'}, \mathbf{s}_{n'}),
\end{aligned}
\label{Eq11}
\end{equation}
where
\begin{equation}
\beta^{n'}_{j_{n}}(\mathbf{r}_{n'}, \mathbf{s}_{n'}) = \prod \limits_{k_{n'}=1}^{N_{n'}} z^{n'}_{j_{n}}(\mathbf{R}_{n'})^{1-s_{k_{n'}}} (1-z^{n'}_{j_{n}}(\mathbf{R}_{n'}))^{s_{k_{n'}}}.
\label{Eq12}
\end{equation}
Eq.~(\ref{Eq11}) and~(\ref{Eq12}) indicate that the output $y^{n'}_{j_{n'}}(\mathbf{r}_{n'})$ moves beyond the multiliner form, becoming a highly nonlinear function of the original input $x(\mathbf{R}_1)$, thus able to capture even higher-order pixel correlations within a larger receptive field spanned by all convolution layers. Hence, the expressivity increases largely with the number of layers. This builds the third theoretical foundation for the potential advantage of TACNN.
\section{Numerical Experiments}
\label{sec:experiments}
\subsection{Method and Setup}
\label{subsec:method}
In all the numerical experiments (i.e. image classification) throughout this work, the images are of size $28\times28$, and we employ $3\times3$ kernels ($N=9$) with stride 1 for every convolution layer. Each tensor kernel is thus a state in a Hilbert space of dimension $2^9=512$. 
Since in our TACNN the convolution kernels are generic tensors, each of which represents an arbitrary quantum state in the entire Hilbert space and they are able to encode fully correlated structures that may lie beyond the representational capacity of TN architectures with fixed bond dimensions.
%
This maximizes the per-kernel expressivity with a manageable number of parameters, thereby making generic tensor a desirable choice in machine-learning tasks such as image classification, where parameter redundancy often leads to overfitting and poorer generalization.
All image‑classification experiments reported here were implemented with PyTorch and run on NVIDIA GPUs.
%
%
In our numerical experiments, except for the standard normalization that renders all the pixel values in $[0,1]$, there was no other data pre-processing and no data augmentation applied prior to transforming the inputs into product states~(Eq.~(\ref{Eq3})). 
For both CNN and TACNN, we adopted Adam optimizer with a learning rate of $2\times10^{-4}$ and a batch size of $100$. We chose cross-entropy loss as the objective function for optimization. All data were obtained and processed from the results of the same $5$ seeds. 
For TACNN, in the cases of one and two tensor convolution layers, we took the best test accuracies from the total training epochs $400$ and $800$, respectively, and then calculated the mean values and standard deviations accordingly. For the CNN model, we adopted the same protocol while restricting the total training duration to $400$ epochs.
%
%
\subsection{Fashion-MNIST Dataset}
\label{subsec:fmnist}
\begin{figure}[tbp]
\centering
\includegraphics[width=1.0\linewidth]{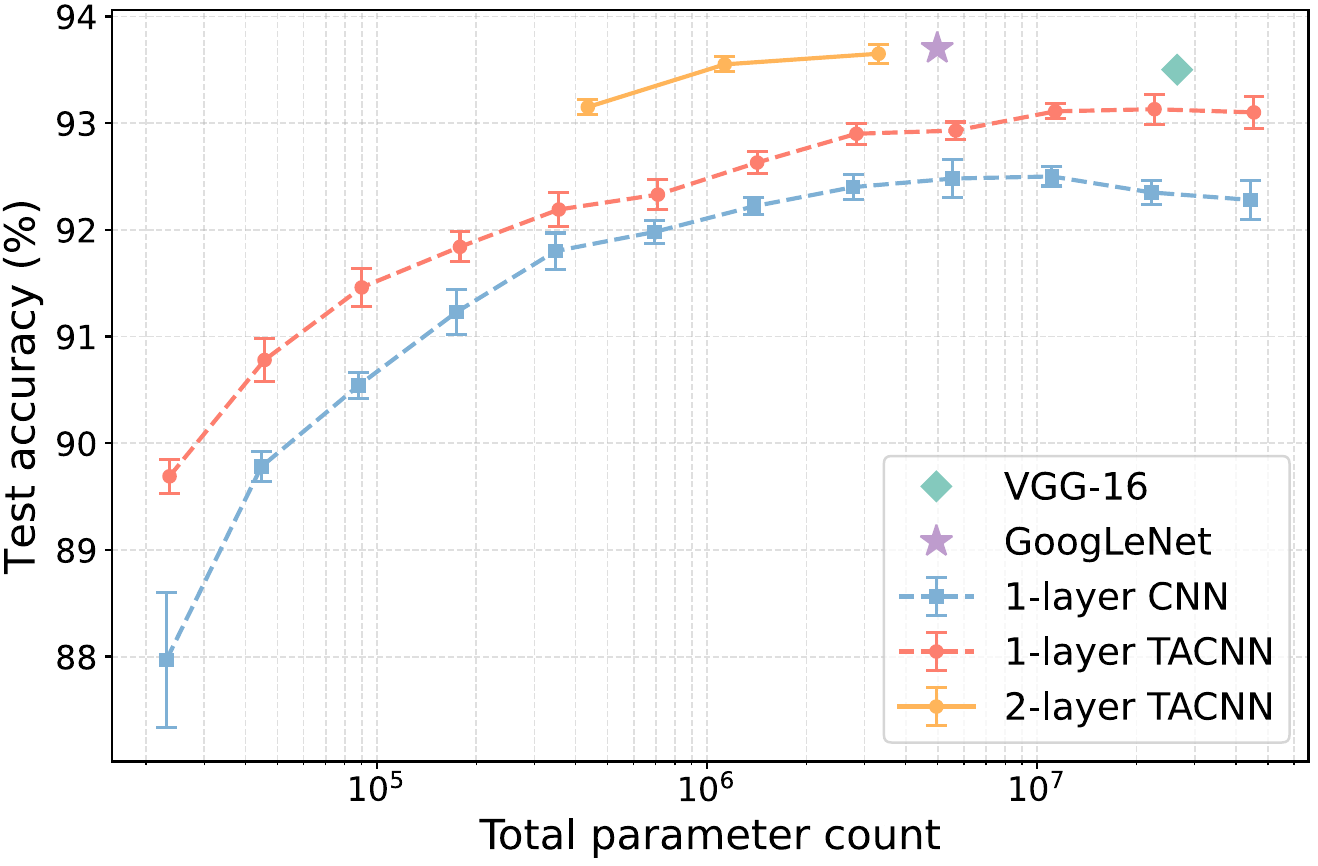}
\caption{Test accuracies of the Fashion-MNIST dataset given by TACNN~(red) and comparable CNN~(blue) with one convolution layer for varying kernel count $2^m$, where $m = 0, 1, \ldots, 11$. For TACNN with two convolution layers~(orange) the kernel counts of the second layer are $16\times16$, $32\times32$ and $64\times64$. Comparison with very deep CNNs VGG-16~(green) and GoogLeNet~(purple) given by Ref.~\cite{xiao2017fashionmnist} further demonstrates the supremacy of TACNN in terms of parameter efficiency.}
\label{Fig2}
\end{figure}
We now proceed to benchmark TACNN with standard image‑classification tasks. Although traditionally MNIST has been used as a baseline dataset for evaluating classification models, it is now widely regarded as a basic sanity check, with the community increasingly inclined to the more challenging Fashion‑MNIST dataset~\cite{xiao2017fashionmnist} for meaningful assessment. Fashion‑MNIST is composed of $70,000$ grayscale images of size $28\times28$, split into $60,000$ training samples and $10,000$ test samples.
In contrast to MNIST, Fashion‑MNIST comprises images with markedly higher visual complexity and is generally viewed as one of the most demanding benchmarks within the MNIST family.
Even conventional deep CNNs such as VGG-16 (Vanilla) and GoogLeNet implemented without data augmentation (DA) can only achieve test accuracies as high as $93.5\%$ and $93.7\%$, respectively~\cite{xiao2017fashionmnist}. Reaching beyond those numbers usually requires more advanced deep CNNs such as ResNet-18 and DenseNet-BC with DA~\cite{xiao2017fashionmnist}, both of which include skip connections. 
To highlight the advantages of our proposed model, we first compare the test accuracies of 1-layer TACNN and CNN (i.e. those with one convolution layer) with the same kernel count $2^m$ where $m = 0, 1, \ldots, 11$. 
Hereafter, we denote the number of kernels for CNN by $N_{\text{CK}}$. Then we compare the test accuracies of 2-layer TACNN (i.e. that with two convolution layers) to the results of conventional deep CNNs without DA as well as TN-based models in the literature. In our experiments, there are three different kernel counts used by the second tensor‑convolution layer: $16\times16$, $32\times32$, and $64\times64$.
In the case of one convolution layer, TACNN consistently outperforms CNN across all number of kernels, as shown in Fig.~\ref{Fig2}. The advantage is particularly significant in the few-kernel regime, where the kernel count ranges from $1$ to $8$. The gap is largest in the $1$-kernel case and narrows with increasing number of kernels. 
Note that in the extreme case of one kernel, TACNN not only beats CNN by a wide margin, but also demonstrates numerical stability far superior to CNN, which has a standard deviation as large as $0.6\%$. This implies that a single tensor kernel is sufficient to effectively capture feature diversity while a single CNN kernel struggles.
Moreover, a rough comparison shows that for CNN to achieve accuracy comparable to that of TACNN, it requires $N_{\text{CK}} \gtrsim 2N_{\text{TK}}$ when $N_{\text{TK}} = 1, 2, 4$, and $N_{\text{CK}} \gtrsim 4N_{\text{TK}}$ when $N_{\text{TK}} = 8, 16, 32$. 
For $N_{\text{TK}} \geq 64$, CNN can no longer match TACNN in accuracy. While the accuracy of CNN saturates at $N_\text{CK}=256$ with $92.5\%$, that of TACNN continues to increase for $N_{\text{TK}} \geq 64$ and saturates at $N_{\text{TK}} = 512$ with $93.1\%$.
Together, these observations quantitatively verifies the stronger per-kernel expressivity of TACNN suggested by the theories in Section~\ref{sec:architectures}.
Note that the increasingly larger gap for $N_{\text{CK}} \geq 512$ is due to overfitting in CNN that leads to decreasing test accuracy despite the far fewer per-kernel parameters. TACNN, on the other hand, does not exhibit overfitting despite having $512$ parameters per kernel while CNN only has $9$ per kernel.
A possible explanation is that for $N_{\text{TK}} \geq 512$, the kernel states after optimization might not yet be able to span a complete $512$-d Hilbert space. Therefore, excessive kernels can still be effective feature extractors without incurring parameter redundancy that causes poorer generalization as observed in CNN.
It is also notable that Table.~\ref{table1} and \ref{table2} show that both 1-layer TACNN with around $32$ to $64$ kernels and 1-layer CNN with $256$ kernels outperform all TN-based models in the literature. Although TN-based models share a similar quantum embedding mechanism and yield an even higher-order multilinear form, capturing feature correlations globally is evidently much less effective than doing so in a local fashion.

\begin{table}[tbp]
\normalsize
\begin{tabular*}{0.45\textwidth}{@{\extracolsep{\fill}}llr}
\hline\hline \\ [-8pt]
Model &  Kernel count  & Test accuracy \\ [2pt]
\hline \\ [-8pt]
1-layer TACNN  & \;\:\:\: 1 & 89.7$\%$ 
               \;\;\;\; \\ [2pt]
               & \;\:\:\: 2 & 90.8$\%$ 
               \;\;\;\; \\ [2pt]
               & \;\:\:\: 4 & 91.5$\%$ 
               \;\;\;\; \\ [2pt]
               & \;\:\:\: 8 & 91.8$\%$ 
               \;\;\;\; \\ [2pt]
               & \;\:\:\: 16 & 92.2$\%$ 
               \;\;\;\; \\ [2pt]
               & \;\:\:\: 32 & 92.3$\%$
               \;\;\;\; \\ [2pt]
               & \;\:\:\: 64 & 92.6$\%$ 
               \;\;\;\; \\ [2pt]
               & \;\:\:\: 128 & 92.9$\%$ 
               \;\;\;\; \\ [2pt]
               & \;\:\:\: 256 & 92.9$\%$ 
               \;\;\;\; \\ [2pt]
               & \;\:\:\: 512 & 93.1$\%$ 
               \;\;\;\; \\ [2pt]
               & \;\:\:\: 1024 & 93.1$\%$ 
               \;\;\;\; \\ [2pt]
               & \;\:\:\: 2048 & 93.1$\%$ 
               \;\;\;\; \\ [2pt]
\hline\hline \\ [-8pt]
2-layer TACNN  & \;\:\:\: 16$\times$16 & 93.2$\%$ 
               \;\;\;\; \\ [2pt]
               & \;\:\:\: 32$\times$32 & 93.6$\%$ 
               \;\;\;\; \\ [2pt]
               & \;\:\:\: \textbf{64$\times$64} & \textbf{93.7}$\%$ 
               \;\;\;\; \\ [2pt]
\hline\hline
\end{tabular*}
\caption{Test accuracies of the Fashion-MNIST dataset given by TACNN architectures with one and two convolution layers for varying number of kernels. The 1‑layer model reaches a maximum accuracy of 93.1$\%$, whereas the 2‑layer model further improves the performance to 93.7$\%$ when applying $64\times64$ kernels in the second layer, representing the highest accuracy achieved in our experiments.}
\label{table1}
\end{table}

In terms of parameter efficiency, 1-layer TACNN is also superior to 1-layer CNN across all kernel counts, as shown in Fig.~\ref{Fig2}. It is clear that to achieve comparable accuracy, TACNN requires fewer parameters than CNN. While in the convolution layer TACNN has exponentially more parameters than CNN, the fully-connected~(FC) layer in both architectures dominates the parameter count, making CNN and TACNN almost indistinguishable in total parameter count, resulting in TACNN still being more efficient.
More specifically, from the flattened input given by the convolution layer to the output of the probabilities of $10$ classes, it is essential to add one or more hidden layers, each followed by an activation function such as ReLU, for the FC-layer to become a nonlinear function approximator capable of learning the probability distributions of different classes in the high dimensional space.
Across all numerical experiments in this work, there is only one hidden layer of $128$ neurons for all architectures. The addition of a hidden layer is the root cause of the FC-layer being the dominant part in the total number of parameters.
In short, although TACNN has exponentially more parameters in the convolution layer, the proper machine-learning practice of adding hidden layer provides an offset, ultimately leading to TACNN being more parameter-efficient than CNN.

\begin{table}[tbp]
\normalsize
\begin{tabular*}{0.45\textwidth}{@{\extracolsep{\fill}}lr}
\hline\hline \\ [-8pt]
\quad Model & Test accuracy \;\; \\ [2pt]
\hline \\ [-8pt]
\quad MPS~\cite{Efthymiou2019} & 88.0$\%$ 
\quad\;\:\:\: \\ [2pt]
\quad PEPS~\cite{Cheng2021} & 88.3$\%$ 
\quad\;\:\:\: \\ [2pt]
\quad EPS + SBS~\cite{Glasser2018} & 88.6$\%$ 
\quad\;\:\:\: \\ [2pt]
\quad MPS + TTN~\cite{Stoudenmire2018} & 89.0$\%$ 
\quad\;\:\:\: \\ [2pt]
\quad Snake-SBS~\cite{Glasser2018} & 89.2$\%$ 
\quad\;\:\:\: \\ [2pt]
\quad LoTeNet~\cite{Arbel2020} & 89.5$\%$ 
\quad\;\:\:\: \\ [2pt]
\quad K-SVM~\cite{xiao2017fashionmnist} & 89.7$\%$ 
\quad\;\:\:\: \\ [2pt]
\quad XGBoost~\cite{xiao2017fashionmnist} & 89.8$\%$ 
\quad\;\:\:\: \\ [2pt]
\quad AlexNet~\cite{xiao2017fashionmnist} & 89.9$\%$ 
\quad\;\:\:\: \\ [2pt]
\quad Low-rank TTN~\cite{Chen2024} & 90.3$\%$ 
\quad\;\:\:\: \\ [2pt]
\quad Residual MPS~\cite{Meng2023} & 91.5$\%$ 
\quad\;\:\:\: \\ [2pt]
\quad Deep TTN~\cite{Nie2025} & 92.4$\%$ 
\quad\;\:\:\: \\ [2pt]
\quad \textbf{1-layer TACNN} & \textbf{93.1}$\%$ 
\quad\;\:\:\: \\ [2pt]
\quad VGG-16~\cite{xiao2017fashionmnist} & 93.5$\%$ 
\quad\;\:\:\: \\ [2pt]
\quad \textbf{2-layer TACNN} & \textbf{93.7}$\%$ 
\quad\;\:\:\: \\ [2pt]
\quad GoogLeNet~\cite{xiao2017fashionmnist} & 93.7$\%$   
\quad\;\:\:\: \\ [2pt]
\hline\hline
\end{tabular*}
\caption{Test accuracies of various models on the Fashion‑MNIST dataset. K-SVM and XGBoost are two of the most commonly used traditional machine-learning methods prior to the paradigm shift to deep neural networks. AlexNet and VGG-16 are typical Vanilla CNNs, whereas VGG‑16 and GoogLeNet represent CNNs with very deep network structures. The rest of the data shown here were obtained by TN-based machine-learning methods. All the abbreviations follow the conventions used in the references listed behind.}
\label{table2}
\end{table}

Moving further by adding a second tensor convolution layer with the scale-up scheme as formulated in Section~\ref{subsubsec:Multitacnn}, we first found that 2-layer TACNN with $16\times16$ kernels attains an accuracy of $93.2\%$, slightly better than the best number of 1-layer TACNN, while the required number of parameters is fewer by more than one order of magnitude.
Such a difference indicates that compared to increasing kernel numbers, increasing depth is a more efficient and effective way to improve the performance. This means that our physically-guided architecture respects the principles of deep neural networks and possesses an inductive bias aligned well with CNNs.
The efficiency is even more pronounced in the case of $32\times32$ kernels, where the 2-layer TACNN achieves $93.6\%$, an accuracy comparable to those given by VGG-16 ($93.5\%$) and GoogLeNet ($93.7\%$), which leverage $23.5\text{x}$ and $4.4\text{x}$ more parameters, respectively~\cite{xiao2017fashionmnist}.
The fact that VGG-16, the best-performing Vanilla CNN, requires such a staggering amount of parameters implies the limitation of Vanilla CNNs' architecture. In contrast,  GoogLeNet further improves the efficiency by introducing a multi-branch Inception design.
Since TACNN follows the design of Vanilla CNNs, it is fair to state that employing tensor kernels enables the model to surpass the state-of-the-art accuracy of conventional CNNs with a far superior parameter efficiency.
With $64\times64$ kernels, 2-layer TACNN further reaches $93.7\%$, being on par with GoogLeNet and still more efficient with a $33.6\%$ parameter saving.
More specifically, the averaged accuracy is $93.65\% \pm 0.09\%$, and within the $5$ seeds we adopted, there is even one seed giving a number as high as $93.79\%$, suggesting that 2-layer TACNN has the potential to outperform GoogLeNet.
Since both VGG-16 and GoogLeNet are very deep CNNs, these results substantiate the significantly stronger per-layer expressivity suggested by the theoretical foundation of TACNN.
Together with the richer per-kernel expressivity illustrated previously, TACNN demonstrates notable synergy stemming from combining physically-principled modeling with the standard protocol of CNNs.

\section{Discussion}
\label{sec:discussion}

In the previous sections we have demonstrated, from theoretical perspectives and numerical simulations, that embedding local superposition states into a CNN framework, although not immediately obvious, substantially enhances the representational capacity of the resulting model. 
Essentially, a single tensor kernel effectively functions as a superposition of an entire family of linear filters, each corresponding to a distinct binary‑like configuration within the local patch. 
In contrast, a conventional CNN kernel represents only one such linear pattern. As a result, a tensor kernel possesses exponentially greater expressive capacity, with the capability of capturing intricate pixel correlations that a linear filter cannot. 
Indeed, a fully entangled tensor can approximate an arbitrary mapping from patch-pixel values to an output, whereas a linear filter is subject to severe structural limitation. 
%
%
This expansion of the accessible function space enables the network to capture significantly richer correlations than classical kernels of the same spatial extent, whose effective vector‑space dimension is limited to the kernel size itself.
We now turn to the fundamental factors that give rise to the markedly different behaviors observed between TACNN and classifiers built on TN architectures. A useful starting point is the observation that fully two‑dimensional TN ansätze, such as the projected entangled-pair state~(PEPS), typically outperforms quasi‑one‑dimensional structures such as matrix product state or density‑matrix-renormalization‑group models when applied directly to intrinsically 2-d learning tasks~\cite{Cheng2021,Stoudenmire2012}. 
This advantage arises from the ability of 2-d tensor networks to capture rich entanglement patterns and complex spatial correlations that are inherently 2-d. Following the 2-d design of CNNs, TACNN operates only with small localized kernels, so it becomes computationally feasible to employ fully generic higher‑order tensors without imposing any structural constraints. 
This flexibility allows TACNN to explore a substantially larger functional space than conventional tensor‑network architectures, whose expressive power is limited by the specific network topology and bond‑dimension restrictions.
Moreover, treating tensor networks directly as classifiers often leads to a rapid increase in computational complexity, whereas TACNN employs only relatively small tensors, making itself significantly more efficient. 
Previous work has also shown that existing optimization strategies for PEPS in machine‑learning settings can be inadequate~\cite{Cheng2021}, particularly when no guiding Hamiltonian is available, in contrast to typical applications in quantum many‑body physics. 
Taken together, these considerations position TACNN as a more practical and effective approach for incorporating tensor‑based methods into machine‑learning models.

In addition to the challenges mentioned above, previous work has shown that quantum convolutional neural network (QCNN) architectures can achieve high accuracy in quantum phase recognition~\cite{Cong2019}. 
This suggests that quantum circuits and their classical approximations via tensor‑network representations might be inherently better suited for capturing long‑range correlations. In contrast, for classical tasks such as image classification, architectures based on local feature extraction, such as conventional CNNs, are likely more favorable.
Overall, these observations demonstrate that the limitations of TN architectures in standard machine-learning tasks stand in sharp contrast to the flexibility of TACNN, whose kernels are fully generic tensors unconstrained by network topology or bond‑dimension restrictions. 
Still, developing improved optimization schemes for tensor networks tailored to machine-learning tasks, particularly in the absence of a target Hamiltonian, remains an important open question for achieving further compression of information, and we leave this direction for future investigation.

\section{Conclusion}
\label{sec:conclusion}

In this work, we investigate a quantum‑inspired convolutional architecture, TACNN, which embeds small quantum states directly into convolution kernels and systematically evaluate its performance on a standard image‑classification benchmark dataset.
Our results demonstrate that TACNN provides a clear advantage over conventional CNNs: it surpasses VGG‑16 and achieves accuracy comparable to GoogLeNet on Fashion‑MNIST, while requiring significantly fewer variational parameters. 
Apart from this, TACNN differs fundamentally from TN-based classifiers. Whereas TN models are constrained by bond dimensions, area‑law entanglement limits, and challenging optimization landscapes, TACNN employs fully generic higher‑order tensors as kernels, thereby having substantially greater expressive capacity, enabling more efficient feature extraction. 
The combination of strong empirical performance, superior parameter efficiency, and direct interpretability through its tensor structure positions TACNN as a powerful and principled framework for advancing physically-guided, explainable deep learning in computer vision.

It is important to note that, unlike the conventional QCNN architecture that relies on a sizable parametrized quantum circuit for each convolution layer~\cite{Cong2019}, our approach replaces only the convolution kernels with small quantum states represented as generic tensors. 
This design circumvents the large circuit depth and noise accumulation characteristic of QCNNs by restricting quantum involvement to the preparation of quantum states represented by shallow, few-qubit circuits. 
Because the kernels in our framework correspond to quantum states with a relatively small register, they could be generated with circuits of depth in line with current coherence times. 
In particular, the reduced number of entangling gates can help suppress error propagation and mitigate decoherence effects, thus potentially allowing for high‑fidelity state preparation even on noisy quantum processors. 
As a result, our architecture is inherently less subject to the constraints of noisy intermediate‑scale quantum~(NISQ) devices~\cite{Preskill2018, Bharti2022}, offering a physically realistic and experimentally accessible pathway for hybrid quantum-classical convolutional models, distinct from QCNN approaches that rely on deep, highly entangled circuits beyond the operational regime of near‑term hardwares.
We therefore anticipate that our algorithm could provide a robust framework for quantum embedding and quantum ML, catalyzing further advancements in this emerging direction.


\par\noindent\emph{\textbf{Acknowledgments}} ---
The authors acknowledge the fruitful discussion with Ying-Jer Kao, Naoki Kawashima, and Chia-Min Chung.
W.-L.T. is supported by the Center of Innovation for Sustainable Quantum AI (JST Grant Number JPMJPF2221) and JSPS KAKENHI Grant Number JP25H01545 and JP26K17054.

\bibliography{draft}

@article{Zhao2024,
  title = {A review of convolutional neural networks in computer vision},
  author = {Zhao, Xia and Wang, Limin and Zhang, Yufei and Han, Xuming and Deveci, Muhammet and Parmar, Milan},
  journal = {Artif. Intell. Rev.},
  volume = {57},
  issue = {4},
  pages = {99},
  year = {2024},
  month = {Mar},
  doi = {10.1007/s10462-024-10721-6},
  url = {https://doi.org/10.1007/s10462-024-10721-6}
}

@article{Carrasquilla2021,
  title = {How To Use Neural Networks To Investigate Quantum Many-Body Physics},
  author = {Carrasquilla, Juan and Torlai, Giacomo},
  journal = {PRX Quantum},
  volume = {2},
  issue = {4},
  pages = {040201},
  numpages = {24},
  year = {2021},
  month = {Nov},
  publisher = {American Physical Society},
  doi = {10.1103/PRXQuantum.2.040201},
  url = {https://link.aps.org/doi/10.1103/PRXQuantum.2.040201}
}

@article{Vicentini2022,
	title={{NetKet 3: Machine Learning Toolbox for Many-Body Quantum Systems}},
	author={Filippo Vicentini and Damian Hofmann and Attila Szabó and Dian Wu and Christopher   Roth and Clemens Giuliani and Gabriel Pescia and Jannes Nys and Vladimir Vargas-Calderón and   Nikita Astrakhantsev and Giuseppe Carleo},
	journal={SciPost Phys. Codebases},
	pages={7},
	year={2022},
	publisher={SciPost},
	doi={10.21468/SciPostPhysCodeb.7},
	url={https://scipost.org/10.21468/SciPostPhysCodeb.7},
}

@article{Orus2019,
  author = {{Or{\'u}s}, Rom{\'a}n},
  title = "{Tensor networks for complex quantum systems}",
  journal = {Nat. Rev. Phys.},
  year = 2019,
  month = aug,
  volume = {1},
  number = {9},
  pages = {538-550},
  doi = {10.1038/s42254-019-0086-7}
}

@article{Cirac2021,
  title = {Matrix product states and projected entangled pair states: Concepts, symmetries, theorems},
  author = {Cirac, J. Ignacio and P{\'e}rez-Garc{\'i}a, David and Schuch, Norbert and Verstraete, Frank},
  journal = {Rev. Mod. Phys.},
  volume = {93},
  issue = {4},
  pages = {045003},
  numpages = {65},
  year = {2021},
  month = {Dec},
  publisher = {American Physical Society},
  doi = {10.1103/RevModPhys.93.045003},
  url = {https://link.aps.org/doi/10.1103/RevModPhys.93.045003}
}

@article{Banuls2023,
  doi = {10.1146/annurev-conmatphys-040721-022705},
  url = {https://doi.org/10.1146%2Fannurev-conmatphys-040721-022705},
  year = 2023,
  month = {mar},
  publisher = {Annual Reviews},
  volume = {14},
  number = {1},
  pages = {173--191},
  author = {Mari Carmen Ba{\~n}uls},
  title = {Tensor Network Algorithms: A Route Map},
  journal = {Annual Review of Condensed Matter Physics}
}

@inproceedings{Novikov2015,
 author = {Novikov, Alexander and Podoprikhin, Dmitrii and Osokin, Anton and Vetrov, Dmitry P},
 booktitle = {Advances in Neural Information Processing Systems},
 editor = {C. Cortes and N. Lawrence and D. Lee and M. Sugiyama and R. Garnett},
 pages = {},
 publisher = {Curran Associates, Inc.},
 title = {Tensorizing Neural Networks},
 url = {https://proceedings.neurips.cc/paper_files/paper/2015/file/6855456e2fe46a9d49d3d3af4f57443d-Paper.pdf},
 volume = {28},
 year = {2015}
}

@InProceedings{Cohen2016,
  title = 	 {Convolutional Rectifier Networks as Generalized Tensor Decompositions},
  author = 	 {Cohen, Nadav and Shashua, Amnon},
  booktitle = 	 {Proceedings of The 33rd International Conference on Machine Learning},
  pages = 	 {955--963},
  year = 	 {2016},
  editor = 	 {Balcan, Maria Florina and Weinberger, Kilian Q.},
  volume = 	 {48},
  series = 	 {Proceedings of Machine Learning Research},
  address = 	 {New York, New York, USA},
  month = 	 {20--22 Jun},
  publisher =    {PMLR},
  url = 	 {https://proceedings.mlr.press/v48/cohenb16.html}
}

@inproceedings{Stoudenmire2016,
 author = {Stoudenmire, Edwin and Schwab, David J},
 booktitle = {Advances in Neural Information Processing Systems},
 editor = {D. Lee and M. Sugiyama and U. Luxburg and I. Guyon and R. Garnett},
 pages = {},
 publisher = {Curran Associates, Inc.},
 title = {Supervised Learning with Tensor Networks},
 url = {https://proceedings.neurips.cc/paper_files/paper/2016/file/5314b9674c86e3f9d1ba25ef9bb32895-Paper.pdf},
 volume = {29},
 year = {2016}
}

@article{Stoudenmire2018,
  title={Learning relevant features of data with multi-scale tensor networks},
  author={Stoudenmire, Edwin M},
  journal={Quantum Science and Technology},
  volume={3},
  number={3},
  pages={034003},
  year={2018},
  url = {https://iopscience.iop.org/article/10.1088/2058-9565/aaba1a},
  publisher={IOP Publishing}
}

@article{Han2018,
  title = {Unsupervised Generative Modeling Using Matrix Product States},
  author = {Han, Zhao-Yu and Wang, Jun and Fan, Heng and Wang, Lei and Zhang, Pan},
  journal = {Phys. Rev. X},
  volume = {8},
  issue = {3},
  pages = {031012},
  numpages = {13},
  year = {2018},
  month = {Jul},
  publisher = {American Physical Society},
  doi = {10.1103/PhysRevX.8.031012},
  url = {https://link.aps.org/doi/10.1103/PhysRevX.8.031012}
}

@article{Levine2019,
  title = {Quantum Entanglement in Deep Learning Architectures},
  author = {Levine, Yoav and Sharir, Or and Cohen, Nadav and Shashua, Amnon},
  journal = {Phys. Rev. Lett.},
  volume = {122},
  issue = {6},
  pages = {065301},
  numpages = {7},
  year = {2019},
  month = {Feb},
  publisher = {American Physical Society},
  doi = {10.1103/PhysRevLett.122.065301},
  url = {https://link.aps.org/doi/10.1103/PhysRevLett.122.065301}
}

@article{Liu2019,
doi = {10.1088/1367-2630/ab31ef},
url = {https://doi.org/10.1088/1367-2630/ab31ef},
year = {2019},
month = {jul},
publisher = {IOP Publishing},
volume = {21},
number = {7},
pages = {073059},
author = {Liu, Ding and Ran, Shi-Ju and Wittek, Peter and Peng, Cheng and García, Raul Blázquez and Su, Gang and Lewenstein, Maciej},
title = {Machine learning by unitary tensor network of hierarchical tree structure},
journal = {New Journal of Physics}
}

@inproceedings{Glasser2019,
 author = {Glasser, Ivan and Sweke, Ryan and Pancotti, Nicola and Eisert, Jens and Cirac, Ignacio},
 booktitle = {Advances in Neural Information Processing Systems},
 editor = {H. Wallach and H. Larochelle and A. Beygelzimer and F. d\textquotesingle Alch\'{e}-Buc and E. Fox and R. Garnett},
 pages = {},
 publisher = {Curran Associates, Inc.},
 title = {Expressive power of tensor-network factorizations for probabilistic modeling},
 url = {https://proceedings.neurips.cc/paper_files/paper/2019/file/b86e8d03fe992d1b0e19656875ee557c-Paper.pdf},
 volume = {32},
 year = {2019}
}

@article{Wang2023,
title = {Tensor Networks Meet Neural Networks: A Survey and Future Perspectives},
journal = {arXiv:2302.09019},
year = {2023},
url = {https://arxiv.org/abs/2302.09019},
author = {Maolin Wang and Yu Pan and Zenglin Xu and Guangxi Li and Xiangli Yang and Danilo Mandic and Andrzej Cichocki}
}

@article{Garcia2024,
  author={Garc{\'i}a, Marcos D{\'i}ez and M{\'a}rquez Romero, Antonio},
  journal={IEEE Access}, 
  title={Survey on Computational Applications of Tensor-Network Simulations}, 
  year={2024},
  volume={12},
  number={},
  pages={193212-193228},
  doi={10.1109/ACCESS.2024.3519676}}

@misc{Nie2025,
  title={Deep Tree Tensor Networks for Image Recognition}, 
  author={Chang Nie and Junfang Chen and Yajie Chen},
  year={2025},
  eprint={2502.09928},
  archivePrefix={arXiv},
  primaryClass={cs.CV},
  url={https://arxiv.org/abs/2502.09928}, 
}

@article{Meshkini2020,
  author = {Meshkini, Khatereh and Platos, Jan and Ghassemain, Hassan},
  editor = {Kovalev, Sergey and Tarassov, Valery and Snasel, Vaclav and Sukhanov, Andrey},
  title = {An Analysis of Convolutional Neural Network for Fashion Images Classification (Fashion-MNIST)},
  journal = {Proceedings of the Fourth International Scientific Conference ``Intelligent Information Technologies for Industry'' (IITI'19)},
  year = {2020},
  publisher = {Springer International Publishing},
  address = {Cham},
  pages = {85--95},
  doi = {10.1007/978-3-030-50097-9_10}
}

@article{xiao2017fashionmnist,
  title = {Fashion-MNIST: a Novel Image Dataset for Benchmarking Machine Learning Algorithms}, 
  author = {Han Xiao and Kashif Rasul and Roland Vollgraf},
  journal = {arXiv:1708.07747},
  year = {2017},
  url = {https://arxiv.org/abs/1708.07747}
}

@article{Efthymiou2019,
  title = {TensorNetwork for Machine Learning}, 
  author = {Stavros Efthymiou and Jack Hidary and Stefan Leichenauer},
  journal = {arXiv:1906.06329},
  year = {2019},
  url = {https://arxiv.org/abs/1906.06329}
}

@article{Cheng2021,
  title = {Supervised learning with projected entangled pair states},
  author = {Cheng, Song and Wang, Lei and Zhang, Pan},
  journal = {Phys. Rev. B},
  volume = {103},
  issue = {12},
  pages = {125117},
  numpages = {8},
  year = {2021},
  month = {Mar},
  publisher = {American Physical Society},
  doi = {10.1103/PhysRevB.103.125117},
  url = {https://link.aps.org/doi/10.1103/PhysRevB.103.125117}
}

@article{Glasser2018,
  title = {From probabilistic graphical models to generalized tensor networks for supervised learning}, 
  author = {Ivan Glasser and Nicola Pancotti and J. Ignacio Cirac},
  journal = {arXiv:1806.05964},
  year = {2018},
  url = {https://arxiv.org/abs/1806.05964}
}

@InProceedings{Arbel2020,
  title = 	 {Tensor Networks for Medical Image Classification},
  author =       {Selvan, Raghavendra and Dam, Erik B},
  booktitle = 	 {Proceedings of the Third Conference on Medical Imaging with Deep Learning},
  pages = 	 {721--732},
  year = 	 {2020},
  editor = 	 {Arbel, Tal and Ben Ayed, Ismail and de Bruijne, Marleen and Descoteaux, Maxime and Lombaert, Herve and Pal, Christopher},
  volume = 	 {121},
  series = 	 {Proceedings of Machine Learning Research},
  month = 	 {06--08 Jul},
  publisher =    {PMLR},
  url = 	 {https://proceedings.mlr.press/v121/selvan20a.html}
}

@ARTICLE{Chen2024,
  author={Chen, Hao and Barthel, Thomas},
  journal={IEEE Transactions on Pattern Analysis and Machine Intelligence}, 
  title={Machine Learning With Tree Tensor Networks, CP Rank Constraints, and Tensor Dropout}, 
  year={2024},
  volume={46},
  number={12},
  pages={7825-7832},
  doi={10.1109/TPAMI.2024.3396386}
}

@Article{Meng2023,
	title={{Residual matrix product state for machine learning}},
	author={Ye-Ming Meng and Jing Zhang and Peng Zhang and Chao Gao and Shi-Ju Ran},
	journal={SciPost Phys.},
	volume={14},
	pages={142},
	year={2023},
	publisher={SciPost},
	doi={10.21468/SciPostPhys.14.6.142},
	url={https://scipost.org/10.21468/SciPostPhys.14.6.142},
}

@article{Stoudenmire2012,
   author = "Stoudenmire, E.M. and White, Steven R.",
   title = "Studying Two-Dimensional Systems with the Density Matrix Renormalization Group", 
   journal= "Annual Review of Condensed Matter Physics",
   year = "2012",
   volume = "3",
   number = "Volume 3, 2012",
   pages = "111-128",
   doi = "https://doi.org/10.1146/annurev-conmatphys-020911-125018",
   url = "https://www.annualreviews.org/content/journals/10.1146/annurev-conmatphys-020911-125018",
   publisher = "Annual Reviews",
   issn = "1947-5462",
   type = "Journal Article"
  }

@article{Cong2019,
  title = {Quantum convolutional neural networks},
  author = {Iris Cong and Soonwon Choi and Mikhail D. Lukin},
  journal = {Nat. Phys.},
  volume = {15},
  pages = {1273–1278},
  year = {2019},
  url = {https://www.nature.com/articles/s41567-019-0648-8}
}

@article{Preskill2018,
  doi = {10.22331/q-2018-08-06-79},
  url = {https://doi.org/10.22331/q-2018-08-06-79},
  title = {Quantum {C}omputing in the {NISQ} era and beyond},
  author = {Preskill, John},
  journal = {{Quantum}},
  issn = {2521-327X},
  publisher = {{Verein zur F{\"{o}}rderung des Open Access Publizierens in den Quantenwissenschaften}},
  volume = {2},
  pages = {79},
  month = aug,
  year = {2018}
}

@article{Bharti2022,
  title = {Noisy intermediate-scale quantum algorithms},
  author = {Bharti, Kishor and Cervera-Lierta, Alba and Kyaw, Thi Ha and Haug, Tobias and Alperin-Lea, Sumner and Anand, Abhinav and Degroote, Matthias and Heimonen, Hermanni and Kottmann, Jakob S. and Menke, Tim and Mok, Wai-Keong and Sim, Sukin and Kwek, Leong-Chuan and Aspuru-Guzik, Al\'an},
  journal = {Rev. Mod. Phys.},
  volume = {94},
  issue = {1},
  pages = {015004},
  numpages = {69},
  year = {2022},
  month = {Feb},
  publisher = {American Physical Society},
  doi = {10.1103/RevModPhys.94.015004},
  url = {https://link.aps.org/doi/10.1103/RevModPhys.94.015004}
}

\end{document}